%% file: main.tex
\def\year{2022}\relax
\documentclass[letterpaper]{article} 
\usepackage{aaai22}  
\usepackage{times}  
\usepackage{helvet}  
\usepackage{courier}  
\usepackage[hyphens]{url}  
\usepackage{graphicx} 
\usepackage{natbib}  
\usepackage{caption} 
\DeclareCaptionStyle{ruled}{labelfont=normalfont,labelsep=colon,strut=off} 
\usepackage{microtype}
\usepackage{booktabs} 
\usepackage{subfigure}
\usepackage{bm}
\usepackage{amsfonts}
\usepackage{soul}
\setul{0.5ex}{0.3ex}
\setulcolor{blue}
\usepackage{makecell}
\usepackage{nicefrac}       
\usepackage{booktabs,multirow}

\usepackage{microtype} 
\usepackage{epsfig}
\usepackage{diagbox}

\usepackage{framed}
\usepackage{empheq}
\usepackage{array}
\usepackage{enumitem}
\usepackage{mdwlist}
\usepackage{cleveref}
\usepackage[linesnumbered, ruled]{algorithm2e} 
\usepackage{algpseudocode}

\usepackage{tcolorbox}
\newtcbox{\mybox}[1][red]
  {on line, arc = 0pt, outer arc = 0pt,
    colback = #1!10!white, colframe = #1!50!black,
    boxsep = 0pt, left = 1pt, right = 1pt, top = 2pt, bottom = 2pt,
    boxrule = 0pt, bottomrule = 1pt, toprule = 1pt}

\usepackage{newfloat}
\usepackage{listings}
\title{Building an Efficient and Effective Retrieval-based Dialogue System \\via Mutual Learning}

\author{
    Chongyang Tao\textsuperscript{\rm $^{1}$},
    Jiazhan Feng\textsuperscript{\rm $^{2}$},
    Chang Liu\textsuperscript{\rm $^{2}$},
    Juntao Li\textsuperscript{\rm $^{2}$},
    Xiubo Geng\textsuperscript{\rm $^{1}$},
    Daxin Jiang\textsuperscript{\rm $^1\thanks{Corresponding author}$}
    \\
}
\affiliations{
    
    \textsuperscript{\rm $^1$}Microsoft Corporation, Beijing, China\\
    \textsuperscript{\rm $^2$}Peking University, Beijing, China\\

    \{chotao,xigeng,djiang\}@microsoft.com \{fengjiazhan,liuchang97,lijuntao\}@pku.edu.cn

}

\usepackage{bibentry}

\begin{document}

\maketitle

\begin{abstract}
Establishing retrieval-based dialogue systems that can select appropriate responses from the pre-built index has gained increasing attention from researchers.  For this task,  the adoption of pre-trained language models (such as BERT) has led to remarkable progress in a number of benchmarks. There exist two common approaches, including cross-encoders which perform full attention over the inputs, and bi-encoders that encode the context and response separately.
The former gives considerable improvements in accuracy but is often inapplicable in practice for large-scale retrieval given the cost of the full attention required for each sample at test time. The latter is efficient for billions of indexes but suffers from sub-optimal performance. In this work, we propose to combine the best of both worlds to build a retrieval system. Specifically, we employ a fast bi-encoder to replace the traditional feature-based pre-retrieval model (such as BM25) and set the response re-ranking model as a more complicated architecture (such as cross-encoder). To further improve the effectiveness of our framework, we train the pre-retrieval model and the re-ranking model at the same time via mutual learning, which enables two models to learn from each other throughout the training process.
We conduct experiments on two benchmarks and evaluation results demonstrate the efficiency and effectiveness of our proposed framework.
\end{abstract}

\input{1-introduction}
\input{2-related}

\input{3-model}

\input{4-experiments}

\section{Conclusion}
In this paper, to build an efficient and effective retrieval-based dialogue system, we propose to combine the fast pre-retriever and the smart response selector. Specifically, we employ a fast bi-encoder to replace the traditional feature-based pre-retrieval model and set the response re-ranking model as a more complicated architecture. To further improve the effectiveness of our framework, we train the pre-retrieval model and the re-ranking model at the same time via mutual learning, which enables two models to learn from each other throughout the training process.
Experimental results on two public benchmarks demonstrate the efficiency and effectiveness of our proposed framework.


\bibliography{aaai22}
\clearpage

\end{document}

%% file: 1-introduction.tex
\section{Introduction}

Building a smart human-computer conversation system has been a long-standing goal in the area of artificial intelligence. 
Recent years have witnessed increasing interest in building an open-domain dialogue system through data-driven approaches with advances in deep learning techniques \cite{sutskever2014sequence,vaswani2017attention,devlin-etal-2019-bert} and the availability of a huge amount of human conversations on social media. By selecting a proper response from the pre-built index with information retrieval techniques~\cite{lowe-etal-2015-ubuntu,whang2020domain} or synthesizing a response with text generation techniques \cite{vinyals2015neural,zhang2019recosa}, existing neural models are now able to naturally reply to user prompts. In this paper, we focus on retrieval-based dialogue systems as they can return fluent and informative responses~\cite{tao2019one} and have powered industrial applications such as the social chatbot Microsoft XiaoIce \cite{shum2018eliza} and the virtual assistant Amazon Alexa \cite{ram2018conversational}.

Existing retrieval-based dialogue systems usually follow the two-stage retrieval and ranking paradigm~\cite{wang2013dataset,li2017alime} where the model first retrieves a bundle of response candidates from a pre-built index through a fast {retrieval} model and then selects an appropriate one as the output response with a more sophisticated response selection model. 
While index building and pre-retrieval methods have been well studied in the information retrieval area, they are less studied in the area of dialogue systems. 
Until recently, most of retrieval-based dialogue systems~\cite{li2017alime,shum2018eliza} depend on traditional term-based information retrieval (IR) methods (such as BM25~\cite{robertson2004simple,qiu2017alime} or TF-IDF~\cite{ji2014information}) to perform the fast pre-retrieval.
While these hand-crafted features are efficient, they
fail to capture the semantics of the context and response candidates beyond lexical matching and remain a major performance bottleneck for the task.

Recently, the use of pre-trained language models (such as BERT~\cite{devlin-etal-2019-bert}) for information retrieval has led to remarkable progress in several tasks. Several studies on dense retrieval methods begin to use pre-trained bi-encoder to cast the input query and index entries into dense representations in a vector space and rely on fast maximum inner-product search (MIPS)~\cite{Shrivastava014Asymmetric} to complete the retrieval. These approaches have demonstrated significant retrieval improvements over traditional IR baselines on open-domain question answering~\cite{karpukhin2020dense}. On the other hand, in the response {re-ranking} stage, two approaches are common: bi-encoders~\cite{lowe-etal-2015-ubuntu,henderson2019training} or poly-encoder~\cite{humeau2020poly} encoding the context and response separately and cross-encoders~\cite{whang2020domain} performing full attention over the inputs.
The former is efficient for billions of indices but suffers from sub-optimal performance.
The latter gives considerable improvements in accuracy but is often inapplicable in practice for large-scale retrieval given the cost of the full attention required for each sample at test time. 
Hence, the challenge we consider is the following:
\textit{How to benefit from accurate cross-attention mechanisms while preserving the fast and scalable response matching?}

In this work, instead of configuring new architectures for response {re-ranking}, we investigate how to build an efficient and effective retrieval-based dialogue system by combining the best of both worlds. Specifically, we train a fast bi-encoder architecture to replace the traditional feature-based pre-retrieval model (such as BM25) and perform the response pre-retrieval with the help of MIPS. In the response {re-ranking} stage, we employ a more complicated and powerful architecture (e.g., cross-encoder) (named \emph{response selector}) to {re-rank} a small number of most promising candidates provided by the fast pre-retrieval model. 
To further improve the effectiveness of our overall systems, we also introduce to train the pre-retriever and the response selector at the same time via mutual learning, which enables two models to learn from each other throughout the training process. By combining the fast pre-retriever and smart response selector, our framework can achieve impressive performance while demonstrating acceptable efficiency over the state-of-the-art cross-encoders.

We conduct experiments with two benchmarks, including Ubuntu Dialogue Corpus~\cite{lowe-etal-2015-ubuntu} and the response selection track of Dialog System Technology Challenge 7~\cite{gunasekara2019dstc7}. On both benchmarks, the model is required to select the most appropriate response from a bundle of candidates or the large index. Evaluation results indicate that:
1) we bring consistent improvements over both the fast bi-encoder and smart cross-encoder with the mutual learning method that transfers knowledge from each other during training. 2) our combined two-stage model is significantly better than the existing models, while also performing at a faster and acceptable test speed.

Our contributions in the paper are four-fold:
\begin{itemize}
    \item First exploration of using the dense pre-retrieval method in retrieval-based dialogue systems;
    \item Proposal of combining the efficient pre-retriever and effective response selector for response retrieval;
    \item Proposal of jointly learning the pre-retriever and the response selector with a mutual learning framework;
    \item Empirical verification of the effectiveness and efficiency of the proposed framework and learning approach on two public data sets.
\end{itemize}

%% file: 2-related.tex
\section{Related Works}

\paragraph{Retrieval-based Dialogues}
Early work for retrieval-based chatbots studies single-turn response selection where the input of a matching model is a message-response pair \cite{wang2013dataset,ji2014information,wang2015syntax}. Recently, more attention is drawn to context-response matching for multi-turn response selection. Representative methods include the dual LSTM model~\cite{lowe-etal-2015-ubuntu}, the multi-view matching model~\cite{zhou2016multi}, the sequential matching network (SMN)~\cite{wu2017sequential}, the deep attention matching network (DAM)~\cite{zhou2018multi}, and the multi-hop selector network (MSN)~\cite{yuan2019multi}.

Recently, pre-trained language models~\cite{devlin-etal-2019-bert,liu2020roberta} on large corpus have shown significant benefits for various downstream NLP tasks, and some researchers have tried to apply them on response selection: to exploit BERT to represent each utterance-response pair and fuse these representations to calculate the matching score ~\cite{vig2019comparison}; to treat the context as a long sequence and conduct context-response matching with BERT~\cite{whang2020domain}. During the post-training on dialogue corpus, this model also introduces the next utterance prediction and masked language model tasks borrowed from BERT to incorporate in-domain knowledge for the matching model; to heuristically incorporate speaker-aware embeddings into BERT to promote the capability of context understanding in multi-turn dialogues~\cite{gu2020speaker}.

\vspace{-2mm}
\paragraph{Efficient Information Retrieval}
Existing information retrieval models~\cite{wang2013dataset,qiu2017alime,nogueira2019passage,nogueira2019multi} usually adopt a pipeline method where an efficient first-stage retriever retrieves a small set of candidates from the entire corpus, and then a powerful but slow second-stage ranker reranks them.
However, most of the models rely on traditional lexical-based methods (such as BM25) to perform the first stage of retrieval and the ranking models of different stages are learned separately. Recently, as a promising approach, Dense Retrieval (DR) has been widely used for Ad-hoc retrieval~\cite{zhan2020learning,chang2020pre,luan2021sparse} and open-domain question answering~\cite{lee2019latent,karpukhin2020dense,xiong2020answering} because it is as fast as traditional methods and can achieve impressive performance.
In retrieval-based dialogue, \citet{humeau2020poly} present the Poly-encoder, an architecture with an additional learnt attention mechanism that represents more global features from which to perform self-attention, resulting in performance gains over Bi-encoders and large speed gains over PLM-based models.
Besides, \citet{henderson2020convert} introduce ConveRT which is a compact dual-encoder pretraining architecture for neural response selection. \citet{vakili2020distilling} utilize knowledge distillation to compress the cross-encoder network as a teacher model into the student bi-encoder model. 

To the best of our knowledge, this paper makes the first attempt to use the dense pre-retrieval method in retrieval-based dialogues and combines the efficient pre-retriever and effective response selector for building an effective and efficient response retrieval system. Besides, different from previous single-directional distillation~\cite{vakili2020distilling}, we jointly train the pre-retriever and the response selector with a mutual learning framework~\cite{zhang2018deep}, which is similar to mutual distillation and enables the knowledge transfer from each other.

%% file: 3-model.tex
\begin{figure*}[t!]
  \centering
  \vspace{-3mm}
  \includegraphics[width=0.92\linewidth]{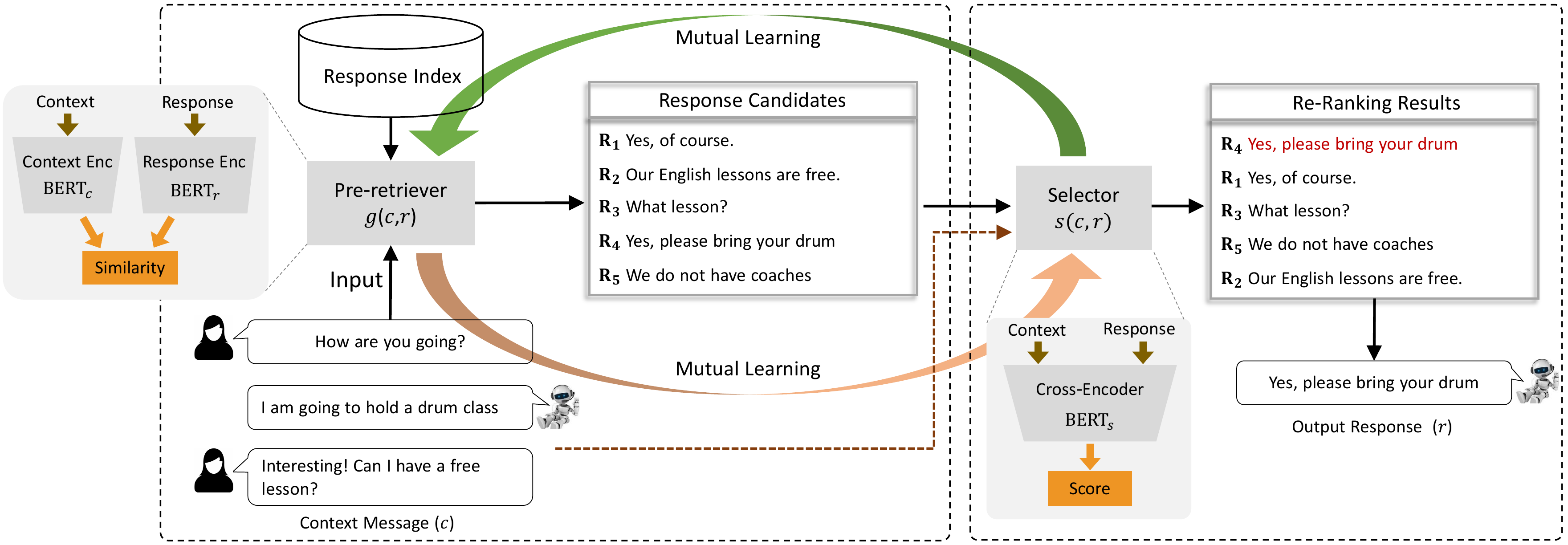}
  \vspace{-2mm}
  \caption{Overall architecture of our model.}
  \vspace{-5mm}
  \label{fig:main-architecture}
\end{figure*}

\section{Problem Formalization}
Given a data set $\mathcal {D} = \{(y,c,r)_z\}_{z=1}^N$ where $c = \{u_1, ...,u_{n_c}\}$ represents a {$n_c$} turns of conversation context with $u_i$ the $i$-th turn, $r$ is a response candidate, and $y \in \{0, 1\}$ denotes a label with $y = 1$ indicating $r$ a proper response for $c$ and otherwise $y = 0$.  The goal of the task of response selection is to build a matching model $\phi(\cdot,\cdot)$ from $\mathcal{D}$. For any input context $c$ and a candidate response $r$, $\phi(c, r)$ gives a score that reflects the matching degree between $c$ and $r$. According to $\phi(c, r)$, one can rank a set of response candidates for response selection. In particular, the definition of $\phi(\cdot,\cdot)$ can be a single-stage model or a two-stage model.

\section{Methodology}

\subsection{Overall Framework}
Retrieval models re-use existing human conversations and select a proper response from a group of candidates for a new user input. Our method is designed within the framework of search engines. Given a message or a conversation context (i.e., a message with several previous turns as conversation history), \emph{the pre-retriever} searches response candidates from an index of existing conversations, and then \emph{the response selector} re-ranks the candidates based on the matching degree between the input and the candidates. 
Specifically, we use a dense retrieval method based on a fast pre-trained bi-encoder architecture as the pre-retrieval model. In the response re-ranking stage, we employ a more complicated and powerful architecture (such as cross-encoder) to re-rank a small number of most promising candidates provided by the fast pre-retrieval model. To further improve the effectiveness of the overall system, we also propose to train the pre-retrieval model and the re-rank model at the same time via mutual learning, which enables two models to learn from each other throughout the training process.


\subsection{Pre-Retriever}
Inspired by the recent dense retrieval~\cite{lee2019latent,zhan2020learning,karpukhin2020dense}, we use a bi-encoder architecture to construct a learnable retriever. The architecture utilizes a separated pre-trained encoder to cast the input context message and index entries into dense representations in a vector space and relies on fast maximum inner-product search (MIPS) to complete the retrieval.
Without loss of generality, we use two BERT~\cite{devlin-etal-2019-bert} models for both encoders, as it is trained on large amounts of unlabelled data and provides strong “universal representations" that can be finetuned on task-specific training data to achieve good performance on downstream tasks.

Specifically, given a context $c = \{u_1,u_2,\ldots,u_{n_c}\}$, where the $t$-th utterance $u_t=\{w_{t,1},\ldots,w_{t,l_t}\}$  is a sequence with $l_t$ words, a response candidate $r = \{r_1,r_2,\ldots,r_{l_r}\}$ consisting of $l_r$ words and a label $y\in\{0, 1\}$,
we first concatenate all utterances in the context as a consecutive token sequence with special tokens separating them, which can be formulated as $x = \{  [\texttt{CLS}], u_1, [\texttt{EOT}], u_2,  [\texttt{EOT}], \ldots,  [\texttt{EOT}], u_{n_c}, $ $ [\texttt{EOT}],  [\texttt{SEP}] \}$. Here $[\texttt{CLS}]$ and $[\texttt{SEP}]$ are the classification symbol and the segment separation symbol of BERT, $[\texttt{EOT}]$ is the "End Of Turn" tag designed for multi-turn context.
For each word of $x$, \emph{token}, \emph{position} and \emph{segment} embeddings of $x$ are summated and fed into pre-trained transformer layer (a.k.a. BERT), giving us the contextualized embedding sequence.
We then project the \texttt{[CLS]} representation to a vector as the context representation following \citet{lee2019latent}. Formally,
\begin{equation}\label{eqn:context-encoder}
\begin{aligned}
v_c = W_c\ \text{BERT}_c \texttt{[CLS]}
\end{aligned}
\end{equation}
where $\text{BERT}_c$ is the context encoder, $W_c$ is the projection matrix for the context \texttt{[CLS]} representation, and $v_c$ is the final context representation containing dialogue history information. We then follow the same scheme to obtain the response representation for a response candidate $r_j$:
\begin{equation}\label{eqn:passage-encoder}
\begin{aligned}
v_r = W_r\ \text{BERT}_r \texttt{[CLS]}
\end{aligned}
\end{equation}
where $\text{BERT}_r$ is the response encoder, $W_r$ is the projection matrix for the response \texttt{[CLS]} representation, and $v_r$ is the final response representation. Finally, the retrieval score is computed as 
\begin{equation}\label{eqn:retriever-score}
\begin{aligned}
g(c_i, r_j) = v_c v_r^\top 
\end{aligned}
\end{equation}

For each training sample, the loss function of the response retriever is defined by
\vspace{-1mm}
\begin{equation} \small
\begin{aligned}
& \mathcal{L}_{\Theta_\mathtt{g}}(c_i, r_i^+, r_{i,1}^-, \ldots, r_{i,\delta_r}^-)  \\
= & -\log(\frac{e^{g(c_i, r_{i}^+)}}{e^{g(c_i ,r_{i}^+)} + \sum_{j=1}^{\delta_r}e^{g(c_i ,r_{i,j}^-)}})
\end{aligned}
\label{eq:pre_loss}
\vspace{-1mm}
\end{equation}
where $r_i^+$ is the true response for a given $c_i$, $r_{i,j}^-$ is the $j$-th negative response candidate randomly sampled from the training set, $\delta_r$ denotes the number of negative response candidate, $\Theta_g$ represents the parameters of the pre-retriever..

\subsection{Response Selector}
To  further  re-rank  a  small  number  of promising candidates provided by the fast pre-retrieval model,
we consider a powerful pre-trained cross-encoder architecture~\cite{devlin-etal-2019-bert} to build the response selector, as it has demonstrated impressive results on various response selection task~\cite{whang2020domain,gu2020speaker}.
Consistent with previous studies \cite{gu2020speaker,whang2020domain}, we also select BERT as the base model for a fair comparison.

Specifically, we first concatenate all utterances in the context as well as the response candidate as a single consecutive token sequence with special tokens separating them formulated as
$x = \{  [\texttt{CLS}], u_1, [\texttt{EOT}], u_2,  [\texttt{EOT}], \ldots,  [\texttt{EOT}], u_{n_c}, $ $ [\texttt{EOT}],  [\texttt{SEP}], r, [\texttt{SEP}] \}$.
Similarly, \emph{token}, \emph{position} and \emph{segment} embeddings are also used. After being processed by BERT$_{s}$, the input sequence is transformed into a contextualized embedding sequence. 
$\text{BERT}_s {\texttt{[CLS]}}$ is an aggregated representation vector that contains the semantic interaction information for the context-response pair.
We then fed $\text{BERT}_s {\texttt{[CLS]}}$ into a multi-layer perception to obtain the final matching score for the context-response pair:
\begin{equation} 
\label{eq-rm}
\begin{aligned} 
s(c,r) &= \sigma ({W}_{2} \cdot f({W}_1 \text{BERT}_s {\texttt{[CLS]}} + b_1) + b_{2})
\end{aligned}
\end{equation}
where ${W}_{\{1,2\}}$ and $b_{\{1,2\}}$ are trainable parameters, $f(\cdot)$ is a $\mathtt{tanh}$ activation function, $\sigma(\cdot)$ stands for a $\mathtt{sigmoid}$ function.

Finally, the training objective of the \emph{response selector} $\mathcal{L}_{\Theta_\mathtt{s}}(\cdot)$ can also be defined as the negative log-likelihood loss similar to Equation~(\ref{eq:pre_loss}).

\subsection{Mutual Learning}
Traditional supervised method individually trains two models to predict the correct labels for the training samples. To improve the effectiveness of our overall systems, we train the pre-retriever and the response selector at the same time via mutual learning~\cite{zhang2018deep}, which enables two models to learn or transfers knowledge from each other throughout the training process. Formally, for a batch of training examples $\{{c}_{i}, r_{i,j}\}_{i=1,j=1}^{i=N,j=\delta_r+1}$, the probability that $\langle{c}_{i}, r_{i,m}\rangle$ is a true context-response pair given by the pre-retriever $\Theta_{g}$ is computed as
 \begin{equation}
p_{m} = \frac{exp(g({c_i,r_{i,m}})/\tau)}{\sum_{j=1}^{\delta_r+1} exp(g({c_i, r_{i,j}})/\tau)} 
 \label{eq:pred}
 \end{equation}
where $g({c_i, r_{i,j}})$ is the output logit of response pre-retriever,  $\tau$ is the temperature to soften $g({c_i, r_{i,j}})$. The output probability of response selector can be computed by the similar way and is denoted as $\bm{q}=[q_1, \cdots, q_{\delta_r+1}]$.

To improve the generalization performance of response pre-retriever $\Theta_{g}$,
we utilize response selector $\Theta_{s}$ to provide training experience in the form of its posterior probability $\bm{q}$. We adopt the Kullback Leibler (KL) Divergence~\cite{kullback1997information} to measure the distance of the two network's predictions $\bm{p}$ and $\bm{q}$, which are predicted by $\Theta_{g}$ and $\Theta_{s}$ separately.
Formally, the KL distance from $\boldsymbol{p}$ to $\boldsymbol{q}$  is computed as
 \begin{equation}
 D_{KL}(\boldsymbol{q}\|\boldsymbol{p}) = \sum_{i=1}^{N} \sum_{m=1}^{M} q_{i,m} \log \frac{p_{i,m}}{q_{i,m}}
  \end{equation}
 
The overall loss function  $\mathcal{L}_{\Theta_{g}}$ for response pre-retriever $(\Theta_{g})$ can be re-defined as 
  \begin{equation}
   \label{eq:sum_loss}
\mathcal{L}_{\Theta_{g}} = \mathcal{L}_{\Theta_{g}} + \alpha \cdot D_{KL}(\boldsymbol{q}\|\boldsymbol{p})
 \end{equation}
where $\alpha$ is the weight for the trade-off two losses. Similarly, we can also use the posterior probability of the response pre-retriever $\Theta_{g}$ to provide training experience for the response selector. The objective loss function $\mathcal{L}_{\Theta_{s}}$ for response selector can be re-defined as 
  \begin{equation}
\mathcal{L}_{\Theta_{s}} = \mathcal{L}_{\Theta_{s}} + \alpha \cdot D_{KL}(\boldsymbol{p}\|\boldsymbol{q})
 \end{equation}
 
In this way, the pre-retriever and response selector both learn to correctly predict the true label of training instances (supervised loss $\mathcal{L}_{\Theta_{\{g,s\}}}$) as well as to match the probability estimate of its counterpart (KL mimicry loss). {The mutual learning method can also be regarded as a mutual distillation process that transfers knowledge estimating the relative quality of the response candidates by each other.} Algorithm 1 describes the pseudo-code of our mutual learning method.

\begin{algorithm}[t!]   
\small
     \caption{The proposed mutual learning method} 
     \KwIn{Training set $\mathcal{D}$, pre-retriever parameters $\Theta_{g}$, response selector parameters $\Theta_{s}$, learning rate $\eta$, number of epochs $n_e$, number of iterations $n_k$;} 
        \For{$e=1,2,...,{n_e}$} 
        { 
            \textbf{Shuffle} training set $\mathcal{D}$\;
            \For{$t=1,2,...,{n_k}$}
            {  
                \textbf{Fetch} a batch of training data $\mathcal{B}$\; 
                 \textbf{Compute} predictions $\boldsymbol{p}$ and $\boldsymbol{q}$ by Eq~(\ref{eq:pred})\; 
                \textbf{Compute} the gradient and \textbf{update} $\Theta_{g}$: \\
                  \begin{equation*}
                \Theta_{g} \leftarrow \Theta_{g} + \eta \frac{\partial L_{\Theta_{g}}(\mathcal{B})}{\partial \Theta_{g} }
                 \end{equation*} 
                \textbf{Update} the predictions $\boldsymbol{p}$ of $\mathcal{B}$; \\
                \textbf{Compute} the  gradient and \textbf{update} $\Theta_{s}$:\\
               \begin{equation*}
                \Theta_{s} \leftarrow \Theta_{s} + \eta \frac{\partial L_{\Theta_{s}}(\mathcal{B})}{\partial \Theta_{s} }
                \end{equation*} 
                \textbf{Update} the predictions $\boldsymbol{q}$ of $\mathcal{B}$;
            }
        } 
    \KwOut{$\Theta_g$, $\Theta_s$.} 
\end{algorithm}

\vspace{-1mm}
\subsection{Inference}
After learning models from $\mathcal{D}$, we first rank the response index according to {$g(c,r)$} and then select top $n_r$ response candidates $\{r_1, \ldots, r_{n_r}\}$ for the subsequent response re-ranking process.
Here we consider two strategies to predict the final response.
In the first strategy, we directly use the matching score of the response selector to obtain the final matching score and output the response with the highest score.

In the second strategy, the final matching score is defined as an integration of the score computed by the pre-retriever and response selector,
\begin{equation} 
\hat{s}(c,r) = g(c,r) + s(c,r) 
\label{eq:ensem}
\end{equation}
The intuition of the above operation is that the pre-retriever focuses on coarse-grained semantic matching based on discourse-level representation, while the response selector focuses on fine-grained word-by-word interaction. We hope to make more accurate predictions with the help of both parties.
We compare the two strategies through empirical studies, as will be reported in the next section.

%% file: 4-experiments.tex
\section{Experiments}
We evaluate the proposed method on two benchmark datasets for both single-sgate and two-stage multi-turn response selection tasks.

\begin{table*}[ht!]
	\centering
	\small
	\vspace{-3mm}
	\resizebox{0.88\textwidth}{!}{
	\begin{tabular}{lcccccccc}
		\toprule[1pt]
		\multirow{2}{*}{} &
		\multicolumn{4}{c}{Sub-task1 of DSTC7} &
		\multicolumn{4}{c}{UbuntuV2}\\ 
		\cmidrule(lr){2-5} \cmidrule(l){6-9}
		Model &  hits@1 & hits@10 & hits@50 & MRR& hits@1 & hits@2 & hits@5 & MRR\\
		\midrule[1pt]
	
		DAM~\cite{zhou2018multi} &  34.7 & 66.3 & - & 35.6 & - & -& -& -\\
		ESIM~\cite{chen2019esim}& 64.5 & 90.2 &  99.4 & 73.5 & 73.4 & 86.6& 97.4& 83.5  \\
		IMN~\cite{gu2019interactive} &- & -& -& -&- &  77.1 & 88.6 & 97.9 \\
		Bi-Enc~\cite{humeau2020poly} &70.9 & 90.6 &  - & 78.1 & 83.6 & - &  98.8 & 90.1 \\
		Poly-Enc~\cite{humeau2020poly} &71.2 & 91.5 & - & 78.2& 83.9 & - & 98.8 & 90.3 \\
		Cross-Enc~\cite{humeau2020poly} &71.7 & 92.4 & - & 79.0&  86.5 & - &  99.1 &  91.9 \\
		\midrule
		Bi-Enc (Our implementation)& 67.5	&91.6 & 98.9 & 76.1 &83.1 & 92.7 & 98.8 & 89.9 \\
		Cross-Enc (Our implementation)& 71.2 & 93.2 & 99.2  & 78.8& 86.6 & 94.3 & 99.3 &  92.0\\ \midrule
		
		Bi-Enc (ML)& 72.4$^\circ$ & 93.7$^\circ$ & 99.2$^\circ$ & 80.1$^\circ$  &	85.7$^\circ$ & 93.8$^\circ$ & 99.0$^\circ$ & 91.5$^\circ$  \\
		Cross-Enc (ML) & 73.9$^\circ$ &  93.9$^\circ$ & 99.4$^\circ$  & 81.3$^\circ$ & 87.4$^\circ$ & 94.7$^\circ$  & 99.2$^\circ$ & 92.6$^\circ$  \\
		Ensemble & 75.3$^\star$ & 94.5$^\star$ & 99.4$^\star$ & 82.3$^\star$ & 87.6$^\star$ & 94.8$^\star$ & 99.2$^\star$ & 92.6$^\star$ \\
		
		\bottomrule[1pt]
	\end{tabular}
	}
	\vspace{-2mm}
	\caption{Results on UbuntuV2 and sub-task1 of DSTC7. Numbers marked with $^\circ$ mean that improvement to the original models is statistically significant (t-test, $p<0.05$). Numbers marked with $^\star$ mean that improvement to the state-of-the-art is significant.} 
	\vspace{-3mm}
	\label{tab:main}
\end{table*}

\subsection{Datasets and Evaluation Metrics}

The first dataset is the track 2 of Dialog System Technology Challenge 7 (DSTC7)~\cite{gunasekara2019dstc7}. The dataset is constructed by applying a new disentanglement method~\cite{kummerfeld2018analyzing} to extract conversations from an IRC channel of technical help for the Ubuntu system.  We use the copy shared by \citet{humeau2020poly} where contains about $2$ million context-response pairs for training.
At test time, the systems were provided with conversation histories,
each paired with a set of response candidates that could be the next utterance in the conversation. Systems needed to rank these options.
We test our model on two sub-tasks. For each dialog context in sub-task 1, a candidate pool of $100$ is given and the contestants are expected to select the best next utterance from the given pool. In sub-task 2, one large candidate pool of $120,000$ utterances is shared by validation and testing datasets. The next best utterance should be selected from this large pool of candidate utterances. In both sub-tasks, there are $5,000$ and $1,000$ dialogues for validation and test respectively.

The second dataset is the {Ubuntu Dialogue Corpus (v2.0)}~\cite{lowe-etal-2015-ubuntu}, which consists of multi-turn English dialogues about technical support and is collected from chat logs of the Ubuntu forum.
We use the copy shared of \citet{jia2020multi}, which has $1.6$ million context-response pairs for training, $19,560$ pairs for validation, and $18,920$ pairs for test.
The ratio of positive candidates and negative candidates is $1:9$ in training, validation set, and test set. 

Following \citet{humeau2020poly}, we employ hits$@k$ and Mean Reciprocal Rank (MRR) as evaluation metrics, where hits$@k$ measures the probability of the positive response being ranked in top $k$ positions among candidates.

\subsection{Baselines}
We compare our method on both the traditional multi-turn response selection scenario as well as the two-stage retrieval scenario. In particular, the following state-of-the-art multi-turn response selection models are selected to compare with our results.
\begin{itemize}
	 \item \textbf{DAM}~\cite{zhou2018multi}: the model follows the representation-matching-aggregation paradigm and the representation is obtained with self and cross attention.
	\item \textbf{ESIM}~\cite{chen2019esim}: the model is the modifications and extensions of the original ESIM~\cite{chen2017enhanced} developed for natural language inference. 
	\item \textbf{IMN}~\cite{gu2019interactive}: the model is a hybrid model with sequential characteristics at the matching layer and hierarchical characteristics at the aggregation layer.
	\item \textbf{Bi-Encoder}~\cite{humeau2020poly}: the model is the same as our pre-retriever.
	\item \textbf{Poly-Encoder}~\cite{humeau2020poly}: the model represents the context and response candidates separately and lets the response interact with the context through an improved attention mechanism.  
	\item \textbf{Cross-Encoder}~\cite{humeau2020poly}: the model is the state-of-the-art models based on pre-trained model. 
\end{itemize}

\begin{table*}[ht!]
	\centering
	\resizebox{0.7\textwidth}{!}{
	\begin{tabular}{lccccc}
		\toprule 
		Model &  hits@1 & hits@2 & hits@5 &  MRR & Test (ms/case)\\
		\midrule
		Poly-Enc & 70.9 & 81.1 & 87.0 & 79.2 & 46\\
		Bi-Enc (ML) &	 72.4 & 81.5 & 88.7 & 80.1  & 45 \\
		Cross-Enc (ML) & 73.9 &  83.0 & 90.4  & 81.3 & 188 \\ \midrule
		Bi-Enc (ML) $\longrightarrow$ Cross-Enc (ML) & 73.9 &  82.8 & 90.1  & 80.9 & 64\\
		Bi-Enc (ML) $\longrightarrow$ Ensemble-Enc (ML)& 75.2$^\star$ &  83.7$^\star$  & 90.9$^\star$   & 81.8$^\star$  & 64\\ 
		\bottomrule 
	\end{tabular}
	}
	\vspace{-2mm}
	\caption{Comparison of single-stage models and two-stage models on the sub-task1 of DSTC7 dataset. We set $n_r= 10$ in the two-stage models and report the score of hits@k ($k=1,2,5$) since only 100 response candidates are given for each dialog context. Numbers marked with $^\star$ mean that improvement to the state-of-the-art is statistically significant (t-test, $p<0.05$).}
	\label{tab:cascade}
\end{table*}

\begin{table*}[ht!]
	\centering
	\resizebox{0.95\textwidth}{!}{
	\begin{tabular}{lccccccc}
		\toprule 
		Model &  hits@1 & hits@2 & hits@5 & hits@10 & hits@50 & MRR & Test (ms/case)\\
		\midrule
		BM25 & 1.4 & 2.0 & 4.2 & 6.0 & 11.9 & 10.0 & -\\ 
		Bi-Enc & 8.6 & 12.2 & 18.7 & 23.2 & 38.1 & 13.6 & -\\
		Bi-Enc (ML) & 10.8 & 16.4 & 23.8 & 30.0 & 46.2 & 17.3 & - \\ \midrule
		BM25 $\longrightarrow$ Bi-Enc & 6.9 & 9.6 & 12.4 & 13.6 & 15.8 & 9.3 & 45\\
		BM25 $\longrightarrow$ Poly-Enc & 7.2 & 9.7 & 12.6 & 13.9 & 15.8 & 9.4 & 46\\
		BM25 $\longrightarrow$ Cross-Enc & 8.0 & 10.4 & 13.5 & 14.6 & 15.8 & 10.3 & 188 \\
		BM25 $\longrightarrow$ Bi-Enc (ML) & 8.1 & 10.1 & 12.7 & 13.8 & 15.6 & 10.0 & 45 \\
		BM25 $\longrightarrow$ Cross-Enc (ML) & 8.8 & 11.8 & 13.9 & 15.0 & 15.7 & 11.0 & 188\\ \midrule
		Bi-Enc $\longrightarrow$ Cross-Enc   & 10.9 & 16.1  &23.8 & 30.9 & 44.6 & 17.3 & 188\\ 

        Bi-Enc (ML) $\longrightarrow$ Cross-Enc (ML) & 12.6$^\star$ & 17.4$^\star$ & 25.2$^\star$ & 30.8 & 48.3$^\star$ & 18.8$^\star$ & 188\\
        Bi-Enc (ML) $\longrightarrow$ Ensemble-Enc (ML) & 12.9$^\star$ & 18.6$^\star$ & 25.6$^\star$ & 32.9$^\star$ & 49.2$^\star$ & 19.3$^\star$ & 188 \\
		\bottomrule 
	\end{tabular}
	}
	\vspace{-1.5mm}
	\caption{Evaluation results on task2 of DSTC7 dataset. We set $n_r=100$ in all two-stage models. It is worth noting that the pre-retrieval with faiss library is very fast and we do not report this part of the time. Numbers marked with $^\star$ mean that improvement to the state-of-the-art is statistically significant (t-test,  $p<0.05$).}
	\vspace{-5mm}
	\label{tab:two_stage}
\end{table*}

\vspace{-1mm}
\subsection{Implementation Details} 
Following~\citet{humeau2020poly}, we select English uncased $\text{BERT}_\text{base}$ (110M) pre-trained on reddit corpus\footnote{\url{https://parl.ai/projects/polyencoder/}} as the context-response matching model and implement our model with \emph{transformers} library provided by huggingface\footnote{\url{https://github.com/huggingface/transformers}}. 
The maximum lengths of the context and response are set to 300 and 72.
Intuitively, the last tokens in the context and the previous tokens in the response candidate are more important, so we cut off the previous tokens for the context but do the cut-off in the reverse direction for the response candidate if the sequences are longer than the maximum length.
We choose 8 as the size of mini-batches for training.
We implement the MIPS with Facebook AI Similarity Search library (Faiss)~\footnote{\url{https://github.com/facebookresearch/faiss}}.
During training, we vary $\alpha$ (Equation (\ref{eq:sum_loss})) in $\{0.5, 1,2,3,4,5\}$, and find that $\alpha=1.0$ is the best choice on both data sets.
We set the number of negative response candidates $\delta_r = 32$ during the training\footnote{Noting that our implementation of Bi-Encoder achieves worse performance than original Bi-Encoder because it considers the other batch elements as negative training samples while we fix the negative samples during training.}.
In two-stage retrieval scenario, we test $n_r$ in $\{10, 50, 100, 200, 500, 800\}$ and set $n_r=100$ for the trade off the efficiency and effectiveness. 
The model is optimized using Adam optimizer with a learning rate set as $3e-5$. The learning rate is scheduled by warmup and linear decay. $\tau$ is set as $3$.
A dropout rate of $0.1$ is applied for all linear transformation layers. The gradient clipping threshold is set as $10.0$.
Early stopping on the corresponding validation data is adopted as a regularization strategy. 

\subsection{Evaluation Results}

\paragraph{Results of traditional response selection.}

We first validated the effectiveness of our framework on traditional response selection scenario. {Table \ref{tab:main} report the evaluation results on sub-task1 of DSTC7 and UbuntuV2 where 10 and 100 response candidates are provided for each input context. }
We can observe that the performance of pre-retriever (e.g., Bi-Enc) and response selector (e.g., Cross-Enc) improve on almost all metrics after they are jointly trained with mutual learning, indicating that the effectiveness of mutual learning on the multi-turn response selection task. Notably, mutual-learning brings more significant improvement to the bi-encoder than cross-encoder on both datasets. The results may steam from the fact that cross-encoder (a stronger model) can provide bi-encoder (a weaker model) with more useful knowledge during the mutual learning phase, but less on the contrary.
With mutual learning, a simple bi-encoder even performs better than the original cross-encoder and poly-encoder on both datasets, although the cross-encoder involves more heavy interaction.
Besides, compared with single-stage models, the ensemble of the pre-retriever and response selector achieves consistently better performance over all metrics on two data sets, demonstrating the advantages of combining matching features of different aspects.

\vspace{-1mm}
\paragraph{Results of two-stage response retrieval.}
We further conduct experiments on two-stage response retrieval scenario. Table \ref{tab:cascade} gives the evaluation results of the sub-task1 of DSTC7.
We use the Bi-Enc (ML) as the pre-retriever, and Cross-Enc (ML) or the ensemble method (as formulated in Eq.~(\ref{eq:ensem})) as the response selector, and denote the two models as \emph{Bi-Enc (ML) $\longrightarrow$ Cross-Enc (ML)} and \emph{Bi-Enc (ML) $\longrightarrow$ Ensemble-Enc (ML)} respectively.
We can find that \emph{Bi-Enc (ML) $\longrightarrow$ Cross-Enc (ML)} achieves slightly worse performance than the cross-encoder. The results are rational since not all the correct response is in the top 10 candidates obtained by Bi-Enc (ML).
However, the two-stage models are about three times faster than Cross-Enc (ML) in inference. In addition, our two-stage model with the ensemble selector is significantly better than the baseline method while also performing at an acceptable test speed. 

Table \ref{tab:two_stage} contains the evaluation results of the sub-task2 of DSTC7. In this task, the model is expected to select the best response from a shared candidate pool of $120,000$ responses, which is more challenging. Due to the huge number of indices, we make use of the MIPS to perform the pre-retrieval, and the time spent in this stage is negligible compared with the response selecting stage.
According to the results, we can observe that: 1) Compared with using BM25 as pre-retriever, Bi-encoder can bring consistent and significant improvement to the overall retrieval system on both datasets, indicating the effectiveness of dense retrieval on the response selection task; 2) Mutual learning can improve the performance of both single-stage models (e.g., Bi-Enc vs Bi-Enc (ML)) and two-stage models (e.g., the models in last two rows); 3) By combining the bi-encoder model and smart cross-encoder model, our two-stage retrieval framework can achieve impressive performance while showing reasonable efficiency constraints compared with other baseline methods, and therefore has good practicality.

\subsection{Discussions}

\begin{figure}[t!]
  \centering
  \vspace{-2mm}
  \includegraphics[width=0.9\linewidth]{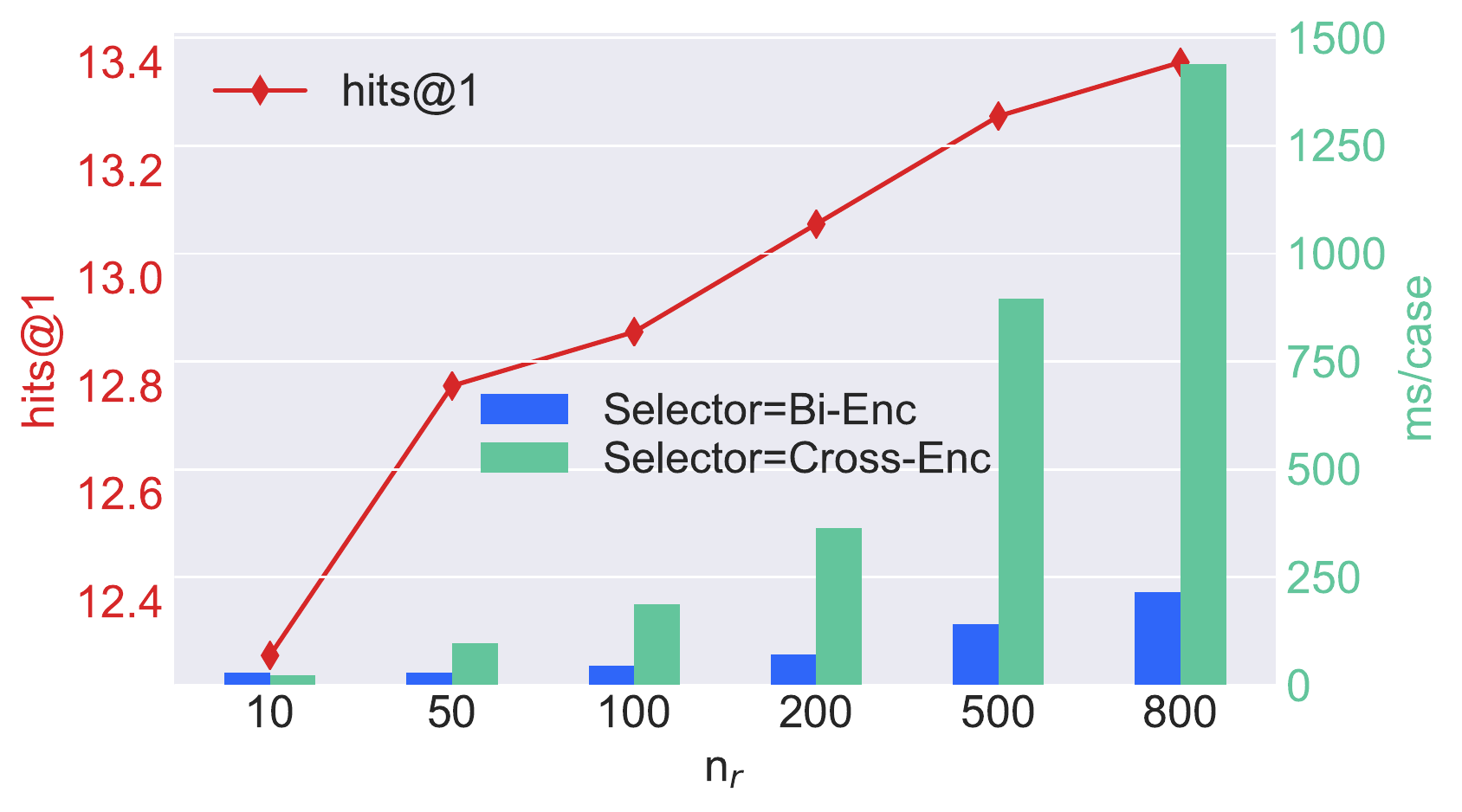}
  \vspace{-4mm}
  \caption{The performance of our two-stage model and average test speed when using the Cross-Enc or Bi-Enc as the selector varies under different $n_r$ on sub-task2 of DSTC7.}
  \vspace{-3mm}
  \label{fig:n_r}
\end{figure}

\paragraph{The impact of $n_r$}
We first check the effectiveness and efficiency of re-ranking performance with respect to the number of top $n_r$ candidates returned from the pre-retriever. Figure~\ref{fig:n_r} illustrates how the hit@1 score of the two-stage model varies under different $n_r$ on sub-task2 of DSTC7 (shown as red lines) and average test speed under different $n_r$ when using the Cross-Enc or Bi-Enc as the selector. We can observe that the retrieval performance increase monotonically as $n_r$ keeps increasing and the improvement becomes smaller when context length reaches $500$. Besides, it can be found that re-ranking as few as $10$ or $50$ candidates out of $120$ thousand from the pre-retriever is enough to obtain good performance under reasonable efficiency constraints.

\begin{figure}[t!]
  \centering
  \includegraphics[width=0.98\linewidth]{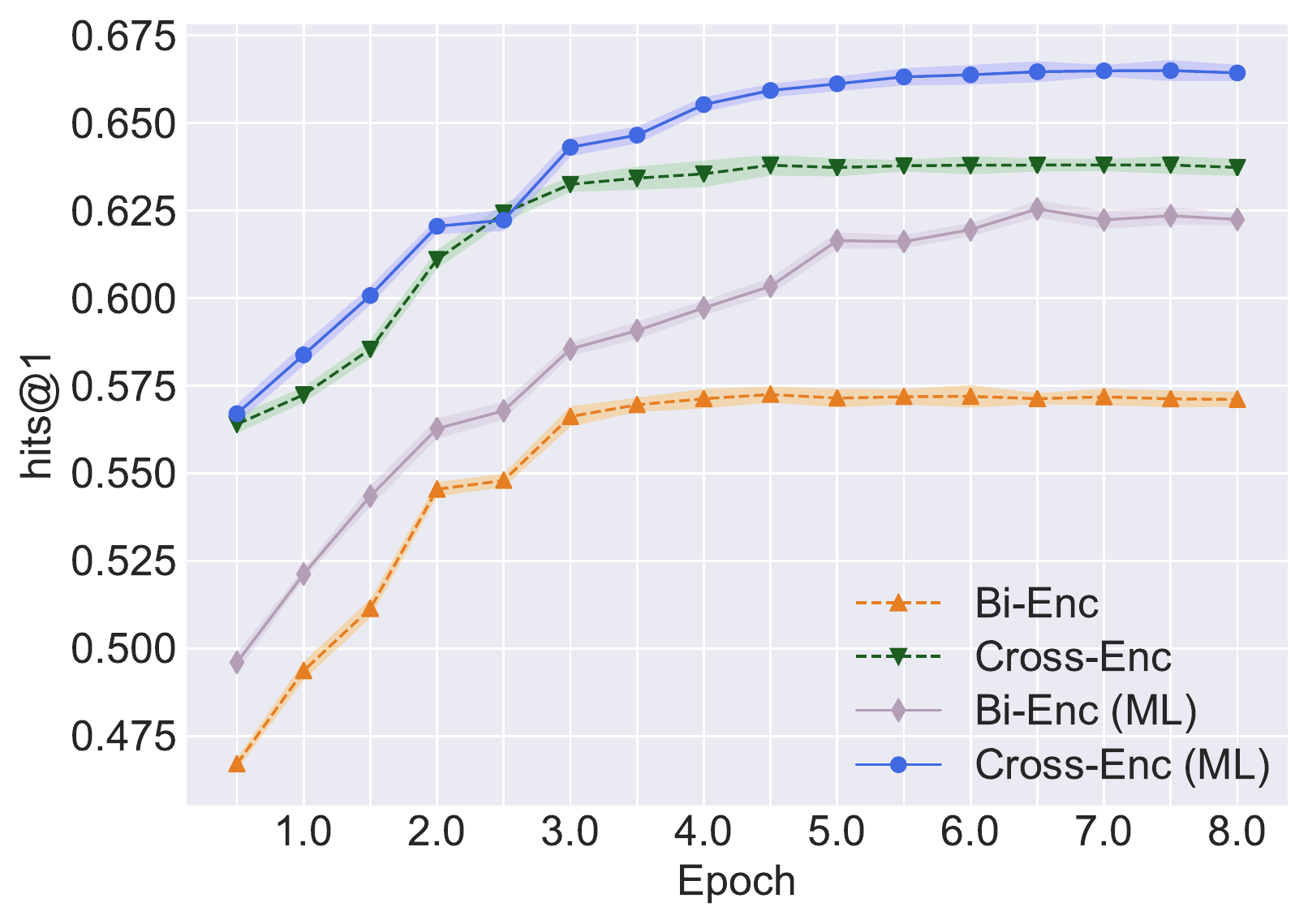}
  \vspace{-4mm}
  \caption{Hits@1 in validation set of various models during the training on sub-task1 of DSTC7. }
  \vspace{-3mm}
  \label{fig:loss}
\end{figure}

\vspace{-1mm}
\paragraph{Training curve of pre-retriever and response selector}
We are curious if the response pre-retriever and response selector can co-improve when they are jointly trained with mutual learning.
Figure~\ref{fig:loss} shows how the hits@1 score of Bi-Encoder,  Cross-Encoder, Bi-Encoder and Cross-Encoder with mutual learning changes with the number of epochs on the validation set of sub-task1 of DSTC7.
From the figure, we can see that jointly training with mutual learning can improve both the performance of pre-retriever (a.k.a, Bi-Enc (ML)) and response selector (a.k.a, Cross-Enc (ML)), and the peer models move at almost the same pace. The results verify our claim that by mutual learning from each other, the two models can get improved together. In addition, we can find that the performance improvement of Bi-Encoder is greater than that of Cross-Encoder. This is because the Cross-encoder can provide Bi-encoder with more useful knowledge during the mutual learning phase.

\begin{figure}[t!]
  \centering
  \includegraphics[width=0.9\linewidth]{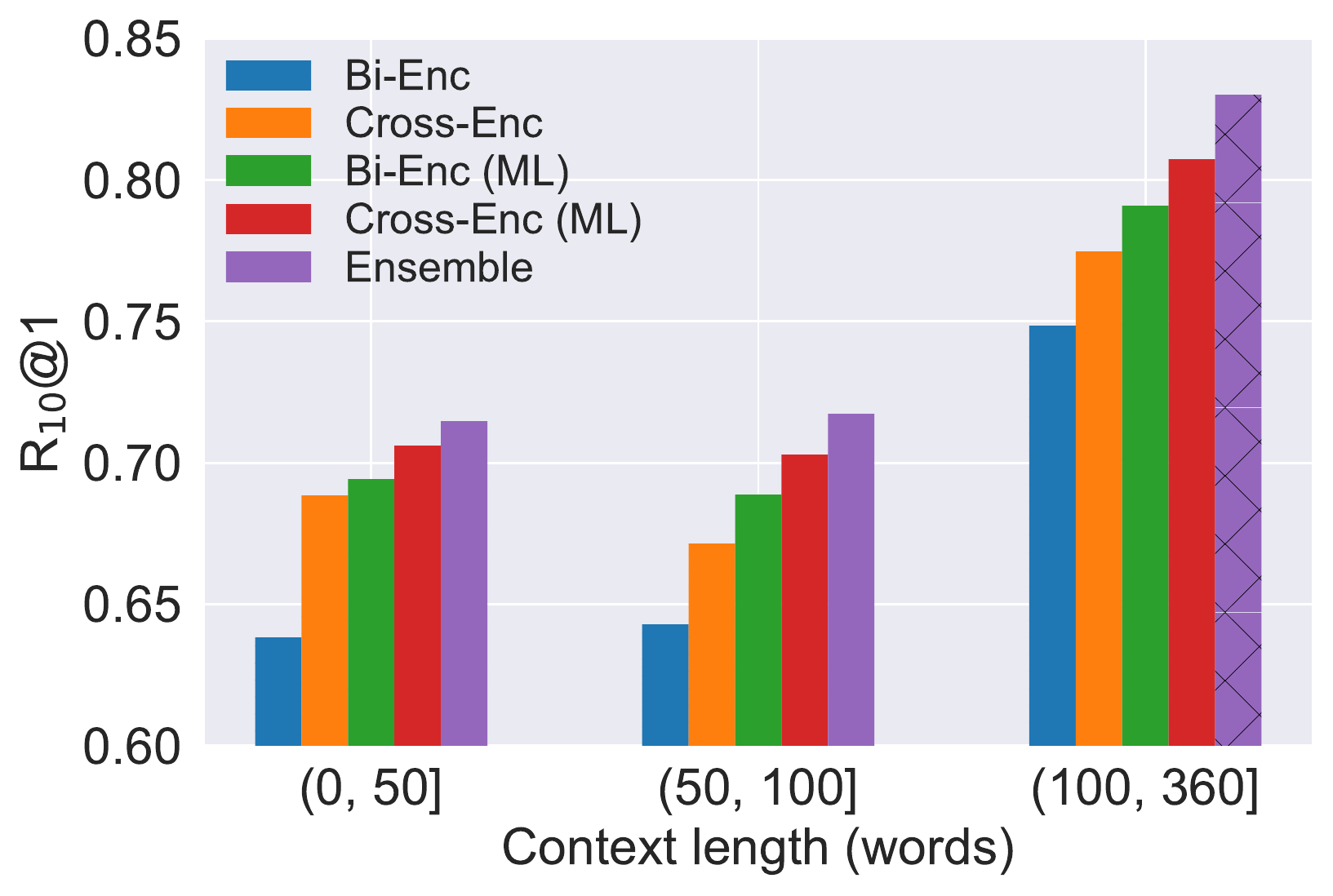}
  \vspace{-3mm}
  \caption{Performance of different models across different length of contexts on sub-task1 of DSTC7. The number of testing samples in the three bins is $339$, $356$, $305$ respectively.}
  \vspace{-4.5mm}
  \label{fig:ctx_len}
\end{figure}

\vspace{-1mm}
\paragraph{The impact of context length}
We further conduct a study to investigate how the length of context influences the performance of these models. Figure~\ref{fig:ctx_len} shows how the performance of the models changes with respect to different lengths of contexts on sub-task1 of DSTC7. We observe a similar trend for all models: they increase monotonically when context length keeps increasing. The phenomenon may come from the fact that the longer context can provide more useful information for response matching. Besides, we can find that mutual learning can bring performance improvements for both the bi-encoder and cross-encoder across all different context lengths, but the improvement is more obvious in the long context (e.g., (100,360]) for cross-encoder and more obvious in the short context (e.g., (0, 50]) for bi-encoder.